\documentclass[11pt]{article}

\usepackage[final]{acl}

\usepackage{times}
\usepackage{latexsym}

\usepackage[T1]{fontenc}
\usepackage[utf8]{inputenc}

\usepackage{microtype}

\usepackage{inconsolata}

\usepackage{graphicx}
\usepackage{booktabs}
\usepackage{amsmath}
\usepackage{pifont}
\usepackage{siunitx}
\usepackage{xspace}
\usepackage{url}

\newcommand{\cmark}{\ding{51}}
\newcommand{\xmark}{\ding{55}}
\newcommand{\both}{$-$}

\newcommand{\benchname}{\textsc{StructuralCost}\xspace}

\title{\benchname: A controlled reading time dataset for modeling human sentence processing difficulty}

\author{
  \textbf{Nina Nusbaumer\textsuperscript{1}},
  \textbf{Iria de-Dios-Flores\textsuperscript{2}},
  \textbf{Corentin Bel\textsuperscript{3, 4}},
\\
  \textbf{Christophe Pallier\textsuperscript{3, 5}},
  \textbf{Guillaume Wisniewski\textsuperscript{1}},
  \textbf{Benoît Crabbé\textsuperscript{1}}
\\
\\
  \textsuperscript{1}LLF, CNRS, Université Paris Cité,
  \textsuperscript{2}COLT, Universitat Pompeu Fabra,
\\
  \textsuperscript{3}Unicog, Neurospin, CEA,
  \textsuperscript{4}LNC2, ENS-PSL,
  \textsuperscript{5}INSERM, CNRS
\\
\\
  \small\texttt{
    \{nina.nusbaumer, guillaume.wisniewski, benoit.crabbe\}@u-pariscite.fr,
    iria.dedios@upf.edu,
  }
\\
  \small\texttt{
    corentin.bel@ens.psl.eu,
    christophe@pallier.org
  }
}

\begin{document}
\maketitle
\begin{abstract}
% We introduce \benchname, a self-paced reading dataset of 475 participants, designed to isolate the processing cost of long dependency resolution. By controlling the distance between the main subject and verb, \benchname~enables systematic measurement of how structural difficulty drives reading times. We show that human reading times at the main verb increase with dependency length and ask whether language model surprisal captures these effects. Using \benchname, we show that existing models fail to replicate the graded difficulty humans exhibit at the main verb as a function of dependency length, despite showing robust and uniform global surprisal-RT fit. This failure is consistent across architectures and model sizes, and suggests that current models capture the predictive component of human processing but not the integration cost that working memory limitations impose. \benchname~provides evaluation data needed to drive progress toward evaluating the cognitive plausibility of language models.

We introduce \benchname, a self-paced reading dataset of 475 participants and \num{40800} observations isolating the processing cost of long-distance subject-verb dependency resolution. We replicate a low-powered psycholinguistic finding at NLP scale, namely that human reading times at the main verb increase with dependency length, driven by syntactic embedding beyond linear distance. Different language models -- spanning $n$-gram models, SSMs, and transformers -- partially mirror this graded difficulty profile, yet underestimate the integration cost humans incur, with a gap that persists across architectures and model sizes. This suggests these models capture the predictive component of human processing but not the full integration cost that working memory imposes. \benchname~provides data needed to drive progress toward evaluating the cognitive plausibility of language models.

\end{abstract}

% Main story: Dependency length affects processing -> but we lack systematic data -> here's our controlled dataset -> here's what it can be used for

% Need to acknowledge / discuss: distinguish natural reading from word by word reading (different encoding and reading strategies): SPR relies more on working memory; transofmers better fit this kind of reading data (anonymous submisision) 

% 480: control for participant instability (eg non detected aphasias)

\section{Introduction}
% Argue why isolating structural effects matters for NLP. Reverse engineering.
% The phenomenon (dependency length matters for processing)
% "Prediction has been proposed as an overarching principle that explains human information processing in language and beyond. To what degree can processing difficulty in syntactically complex sentences – one of the major concerns of psycholinguistics – be explained by predictability, as estimated using computational language models, and operationalized as surprisal (negative log probability)?" (Huang et al., 2024)

An actively debated question in computational psycholinguistics is whether language models (LMs) can serve as computational models of human sentence processing \citep{wilcox2020predictive, oh2023transformer}; essentially, whether the mechanisms driving LMs' text processing mirror those humans deploy during language comprehension. The question has practical stakes: a cognitively plausible LM could be used to test hypotheses at scale that are otherwise difficult or impossible to probe experimentally and to leverage model failures to sharpen sentence processing theories.

Language comprehension is incremental and constrained by two key components: prediction and working memory. As readers encounter each word, they continuously predict what comes next, and processing effort rises when those predictions fail \citep{hale2001probabilistic, levy2008expectation}. At the same time, comprehension relies on working memory, which temporarily hold syntactic information until it can be integrated \citep{gibson2001dependency, lewis2006computational}, e.g., the subject of a sentence remains active in memory until its main verb is later encountered. 
% These two sources of processing difficulty are captured by complementary psycholinguistic theories: \textit{surprisal theory} \citep{hale2001probabilistic, levy2008expectation} attributes reading difficulty to prediction error, while \textit{memory-based accounts} \citep{gibson2001dependency, lewis2006computational} attribute it to the cost of maintaining unresolved dependencies -- a cost that grows with the distance between a head and its dependent.

% LMs assign probabilities to words in context, making their surprisal estimates a natural proxy for the predictive component of human processing. Reading times -- the primary behavioral proxy for online cognitive effort in psycholinguistics -- confirm that LM surprisal correlates reliably with human processing difficulty

LMs assign probabilities to words in context, which are a natural proxy for simulating the predictive component of human language processing. Numerous studies show that the surprisal of models, a simple log-transformation of these probabilities, correlates reliably with reading times -- the primary behavioral proxy for online cognitive effort in psycholinguistics (e.g., \citealp{smith2013effect, goodkind-bicknell-2018-predictive, futrell2021natural}).
% , and a growing body of work asks whether LMs can serve as cognitive models of sentence processing \citep{[refs]}. 
However, surprisal fit alone does not reveal whether LMs capture the memory-based component of language processing, one of its key signatures being the cost of resolving long dependencies: the further apart a syntactic head and its dependent are, the greater the cognitive effort required to resolve that dependency \citep{gibson2001dependency, lewis2006computational}. This is a theoretically important distinction: LM architectures differ greatly in how they handle distance, from $n$-gram models with a hard local window to transformers with unconstrained self-attention. Thus, evaluating whether LMs replicate the graded effect of dependency length is a direct test of cognitive plausibility.

% Such an evaluation requires behavioral data combining experimental control with sufficient scale. Large naturalistic corpora provide scale but confound dependency length with lexical factors (e.g., frequency and length) which independently affect RTs regardless of syntactic structure \citep{oh-etal-2024-frequency}; controlled psycholinguistic experiments isolate structural effects but lack the scale needed for model evaluation. No existing dataset combines both.

Measuring this graded difficulty profile is inherently challenging, as it requires isolating the impact of syntactic structure from confounding lexical factors, such as word frequency and length, which also influence processing times \citep{oh-etal-2024-frequency}. Large naturalistic corpora provide the necessary scale for model evaluation but are often confounded by these lexical variables, making it difficult to separate the effects of dependency length from other influences. On the other hand, controlled psycholinguistic experiments are effective at isolating structural effects but typically lack the scale needed to evaluate models comprehensively. No existing dataset successfully combines both experimental control and large-scale data. To fill this gap, we introduce \benchname, a manually curated dataset of \num{40800} self-paced reading observations from \num{475} participants, including individual working memory scores.\footnote{\benchname is available at \url{https://doi.org/10.34847/nkl.7c467lyv}.  Experiment: \url{https://github.com/NinaNusb/Self-paced-reading-task}. Analysis: \url{https://github.com/NinaNusb/lm-human-processing.git}.} \benchname is built following psycholinguistic standards applied at NLP scale: it holds the main subject-verb pair constant while systematically varying the distance between them, isolating the processing cost of dependency length from lexical confounds. 

Using \benchname, we show that reading times at the main verb increase with dependency length -- an effect driven by syntactic depth beyond linear distance alone -- and that existing LMs only partially mirror this graded difficulty profile despite achieving strong global surprisal-RT fit. Models tend to underestimate the integration cost humans incur, consistently across architectures and model sizes, pointing to a fundamental limitation for treating them as plausible mechanistic models of sentence comprehension: current LMs seem to capture the predictive component of human processing but not the memory-based integration cost. 

Our contributions are threefold:
(1) we release \benchname, a large-scale controlled dataset that isolates dependency length effects on human reading times, replicating at large-scale previously low-powered psycholinguistic findings; \benchname is a controlled diagnostic tool, targeting a specific phenomenon, and not a general benchmark; this limited scope is a design choice, not a limitation. The use of established measures (surprisal-RT alignment, mixed models) is deliberate, so that the results remain directly comparable to existing psycholinguistic and NLP resources, the novelty being in the resource and its controlled manipulation.
(2) In this controlled context, we observe that syntactic structure beyond mere linear distance modulates processing difficulty.
(3) We observe that current LMs, across architectures and sizes, only partially replicate this graded difficulty profile despite strong global surprisal-RT fit, revealing a systematic blind spot in how cognitive plausibility is evaluated. We argue this motivates a shift toward controlled, memory-sensitive datasets as a criterion for cognitive plausibility.

\section{Related Work}

\paragraph{Toward cognitively plausible LMs.}

% Language models -- on their capacities to model human cognition
    % LMs are good candidates for modeling HSP because they are prediction-driven
    % LMs are sometimes good predictors of RTs 
        % ngrams: Michaelov \& Levy (2026)
        % SSMs
        % transformers
    % bigger is not better
        % Michaelov et al 2024 on recurrent models vs transformers
        % Oh & Schuler 2023 on training steps and capacity
    % LMs underestimate HSP difficulty
        % Oh & Linzen 2025
    % Memory capacity as a key factor of RT fit
        % Kuribayashi et al 2022 on context limitations
        % De Varda & marelli 2024 on locally biased transofrmers
        % Timkey & Linzen 2023 on the limited memory capcpity

Because LMs assign probabilities to words in context, their surprisal estimates, defined as $-\log_2 p (\textrm{word} | \textrm{context})$, are a natural proxy for the predictive component of human language processing, and a growing body of work confirms they correlate with human reading times \citep{goodkind-bicknell-2018-predictive, futrell2021natural}. However, recent work finds an inverse scaling trend: larger models do not always yield better psychometric fit \citep{oh2022comparison, oh2023transformer, shain2024large}, likely because state-of-the-art models have become \textit{superhuman} at next-word prediction \citep{oh2025model}, causing their surprisal to underestimate the difficulty humans experience.

A parallel line of work asks whether architectural constraints can bring models closer to human behavior. \citet{devarda2024locally} and \citet{timkey2023language} show that inductive biases, beyond raw prediction accuracy, shape psychometric fit. \citet{kuribayashi2022context} provide a mechanistic explanation: transformer self-attention gives equal weight to all prior context, removing the distance-dependent decay that produces locality effects in human processing. They show that imposing context limitations on transformers partially restores these effects. 
% Similarly, \citet{devarda2024locally} and \citet{timkey2023language} conclude that models with limited or locally biased memory produce surprisal profiles that better approximate human difficulty at structurally demanding positions. 

However, these constrained models have been evaluated primarily on reaction times from naturalistic datasets: Dundee \cite{kennedy2005parafoveal}, Provo \cite{luke2018provo}, MECO \cite{siegelman2022expanding}, UCL \cite{frank2013reading}, the SPR Brown corpus \cite{smith2013effect}, and Natural Stories \cite{futrell2021natural}\footnote{Except \citeauthor{timkey2023language}, who studied disambiguation.}. While these corpora provide valuable insights into the global ability of LMs to fit general behavioral predictability, the uncontrolled variability in naturalistic data -- where frequency, length, and syntactic structure are confounded -- makes these datasets less well suited for modeling processing cost arising from structural complexity. Therefore, to be able to evaluate the cognitive plausibility of LMs, we must move beyond global fit on natural text and adopt controlled datasets that systematically manipulate structural and memory demands.

\begin{table*}[htp]
\centering
\small
\begin{tabular}{llp{12cm}}
\toprule
Condition & Dep. & Example Sentence \\
\midrule
\multicolumn{3}{l}{\textit{Critical sentences}} \\
Baseline      & 0 & The \colorbox{blue!25}{violinist} \colorbox{orange!25}{leaves} the stage before the audience loudly applauds the great performance. \\
PP            & 4 & The \colorbox{blue!25}{violinist} in the large orchestra \colorbox{orange!25}{leaves} the stage before the audience loudly applauds. \\
SRC           & 4 & The \colorbox{blue!25}{violinist} that followed the conductor \colorbox{orange!25}{leaves} the stage before the audience loudly applauds. \\
ORC           & 4 & The \colorbox{blue!25}{violinist} that the conductor followed \colorbox{orange!25}{leaves} the stage before the audience loudly applauds. \\
2$\times$SRC  & 9 & The \colorbox{blue!25}{violinist} that followed the conductor that worked in the orchestra \colorbox{orange!25}{leaves} the stage. \\
2$\times$ORC  & 9 & The \colorbox{blue!25}{violinist} that the conductor that the whole orchestra admired followed \colorbox{orange!25}{leaves} the stage. \\
\midrule
\multicolumn{3}{l}{\textit{Control sentences}} \\
Baseline      & 0 & The \colorbox{blue!25}{waiter} \colorbox{orange!25}{stacks} the plates with the dirty apron next to the kitchen sink. \\
PP-short      & 4 & The \colorbox{blue!25}{waiter} with the dirty apron \colorbox{orange!25}{stacks} the plates next to the kitchen sink. \\
PP-long       & 9 & The \colorbox{blue!25}{waiter} with the dirty apron next to the kitchen sink \colorbox{orange!25}{stacks} the plates. \\
\bottomrule
\end{tabular}
\caption{Example sets of critical and control sentences. Dep. (dependency length) = number of intervening words between the \colorbox{blue!25}{main subject} and \colorbox{orange!25}{main verb}. \textit{Critical sentences} cross syntactic structure (SRC/ORC) with embedding depth (single/double), plus a 0-word baseline and a 4-word linear-distance control (PP). \textit{Control sentences} manipulate linear distance via prepositional phrases, with no syntactic embedding, providing a reference class for dissociating distance from structural complexity.}
\label{tab:stimuli}
\end{table*}

\paragraph{Limitations of existing datasets. } % to model human processing difficulty
    
% Limits of existing datasets to model human processing difficulty
    % large naturalistic datasets are not controlled enough
    % controlled datasets are too small to evaluate language models
    
% Like the current work, Grodner \& Gibson (2005) collected data for ...

Evaluating the cognitive plausibility of LMs involves testing their sensitivity to structural effects alongside their general ability to predict reading times. We identify four criteria that, we argue, datasets must satisfy to support such evaluation: 
\textbf{(C1)~Isolation} -- sentences must appear in isolation to remove the influence of prior context and focus solely on working memory limits; \textbf{(C2)~Controlled variables} -- the dataset must systematically vary dependency length while keeping vocabulary and lexical properties constant, ensuring that any increase in difficulty is due to memory load, not harder words; \textbf{(C3)~Sufficient items} -- a broad and varied set of unique sentences to provide robust model evaluation; \textbf{(C4)~Sufficient participants} -- enough participants per sentence to ensure reliable statistical estimates.
Existing resources are insufficient for the evaluation of LMs' potential sensitivity to structural complexity as neither meets all three criteria: NLP-oriented datasets offer large-scale naturalistic data but lack experimental control to isolate structural effects, while psycholinguistic datasets provide strong control but lack the items and participants needed for model evaluation. \benchname is designed to satisfy all criteria simultaneously (see Appendix~\ref{app:dataset_comparison} for a detailed evaluation of existing datasets against these criteria).

\section{Dataset Creation}
\paragraph{Design rationale.}

Our dataset targets a well-documented prediction from memory-based theories of sentence processing: longer dependencies between a subject and a verb yield higher integration costs at the verb \citep{gibson2001dependency, lewis2006computational}. To isolate this effect, we contrast sentences that have maximally overlapping vocabulary but varying distance between the main subject and main verb, which we manipulate using intervening material.\footnote{We focus on English, whose relative clause and center-embedding structures are well-documented in the psycholinguistic literature, providing robust predictions against which to evaluate model behavior.}

Dependency length can increase in two ways: linearly, by inserting prepositional phrases or also hierarchically, by embedding relative clauses. Both increase subject-verb distance, but differ in processing demands: linear modifications incrementally extend dependencies, while hierarchical embeddings introduce qualitatively greater complexity by nesting clauses and requiring the reader to manage multiple layers of syntactic information, increasing storage cost on working memory \citep{gibson2001dependency, LiuWuLiu+}, though see Section \ref{sec:discussion} for an unexpected reversal of this prediction at double embedding. Structural complexity may also contribute independently of dependency length: object relative clauses are consistently harder than subject ones even at matched distances \citep{gordon2001memory}. Individual differences in working memory capacity (WMC) may further modulate these effects, with lower-WMC readers showing larger slowdowns under long-distance dependencies \citep{king1991individual, just1992capacity}. The controlled structure of our design allows us to tease apart these potentially confounded factors, through two sentence types: \emph{critical sentences} and \emph{control sentences}.

\textit{Critical} sentences are of primary interest: they manipulate syntactic structure while keeping dependency length constant across conditions. The intervening material between subject and verb is either a subject relative clause (SRC) or an object relative clause (ORC), singly embedded (4-word dependency) or doubly embedded (9-word dependency), yielding four core conditions. Two additional conditions anchor the design: a 0-word dependency baseline with no intervening material, and a prepositional phrase (PP) condition at 4 words. The PP condition is critical to the logic of the design: it matches the singly embedded relatives in surface dependency length while introducing no additional clause (i.e., no embedded subject and verb) between subject and verb, isolating linear distance as a factor independently of clausal embedding. Critical items thus comprise six conditions in total (an example set is shown in Table~\ref{tab:stimuli}).

\textit{Control} sentences extend the linear distance manipulation introduced by the PP condition in the critical items. They augment dependency length through prepositional phrases of increasing length (0, 4, and 9 words), without embedding an additional clause; an example set can be seen in Table~\ref{tab:stimuli}. Matched to critical items on dependency length but not hierarchical complexity, they allow direct dissociation of linear distance from structural depth effects, with the PP conditions serving as the common reference point between the two sets.

\paragraph{Stimulus construction.}

Sentences were generated with LLaMA-3.2 \citep{grattafiori2024llama} and Perplexity.ai (\citeyear{perplexity2024}) under tight constraints (identical subject-verb pairs, minimal lexical variation, semantic plausibility), then manually reviewed and extensively corrected for naturalness. Complete construction details (constraints, correction criteria, and item statistics) are provided in Appendix~\ref{app:appendix_datasheet}. LLM output served only as an initial drafting aid to vary semantic domains while preserving structural consistency across conditions; final acceptance required independent manual judgments of plausibility, semantic equivalence, and naturalness across all six condition-mates in a set. The final dataset comprises \num{60} sentences per condition across the six critical item conditions (\num{360} sentences total) and \num{30} sentences per condition across the three control conditions (\num{90} sentences total), for a grand total of \num{450} sentences.\footnote{Sentences were matched in length (\num{13}–\num{15} words) to prevent reading-strategy differences across conditions, since RTs were collected via non-cumulative moving-window self-paced reading. Shorter sentences were extended to match the double-embedded ones by appending a coordinated clause in the post-critical region.}

\paragraph{Data collection.}
Reading times were collected from native English speakers recruited via Prolific, an online crowdsourcing platform widely used in cognitive research, and using a self-paced reading (SPR) procedure. Participants read one list of 90 counterbalanced sentences (details in Appendix \ref{app:appendix_datasheet}); comprehension was verified using manually curated comprehension questions and only participations with a score $\geq$\SI{70}{\percent} were kept, yielding a total of \num{475} participants. Sentences with word reading times below \SI{100}{\milli\second} or above \SI{3000}{\milli\second} are excluded from the dataset as likely reflecting inattention or technical errors (<\SI{5}{\percent} of the data), yielding a final dataset of \num{27103} critical sentences and \num{13663} control sentences. After the reading task, participants performed a working memory capacity test using the operation span test. Full participant details, exclusion criteria, task parameters, participant instructions, consent procedure, and recruitment details are provided in the Datasheet in Appendix~\ref{app:appendix_datasheet}.

\begin{table*}[t]
\centering
\small
\begin{tabular}{lccccccc}
\toprule
Condition & Dep. & N & Mean RT & SD & Median RT & \multicolumn{2}{c}{Cohen's $d$ (vs.\ baseline)} \\
\cmidrule(lr){7-8}
 & (words) & & & & & $d$ & 95\% CI \\
\midrule
Baseline        & 0 & 4,574 & 490 & 308 & 422 & —      & — \\
PP              & 4 & 4,562 & 512 & 321 & 435 & $-0.08$  & $[-0.12, -0.04]$ \\
SRC             & 4 & 4,549 & 539 & 370 & 439 & $-0.15$  & $[-0.19, -0.12]$ \\
ORC             & 4 & 4,496 & 583 & 408 & 467 & $-0.19$  & $[-0.24, -0.15]$ \\
2$\times$SRC    & 9 & 4,545 & 504 & 302 & 436 & $-0.11$ & $[-0.15, -0.08]$ \\
2$\times$ORC    & 9 & 4,377 & 553 & 383 & 438 & $-0.14$ & $[-0.18, -0.10]$ \\
\bottomrule
\end{tabular}
\caption{RTs (ms) at the main verb by condition, with Cohen's $d$ relative to the baseline (0-word dependency).}
\label{tab:rt_by_condition}
\end{table*}

\section{Human Processing Behavior Analysis}
% Before turning to the ability of LMs to predict reading times, we first report a series of validation experiments aimed at assessing the quality of the collected data. In particular, we verify that human reading times exhibit the expected pattern, allowing us to tease apart the respective contributions of linear distance and syntactic structure.

%Figure~\ref{fig:rt_by_position} displays average log-transformed RTs at each word position for all six conditions. A consistent spike at the main verb (in red) is visible across all conditions, with narrow \SI{95}{\percent} CIs despite large-scale online collection, indicating a stable signal. The gray curves (9-word dependencies) rise more gradually than the sharper peaks of shorter structures, reflecting the complexity of doubly-nested structures and successive reanalyses before resolution at the main verb. The expected end-of-sentence rise in the gray curves contrasts with the flatness of shorter structures (blue/orange), where the final position marks the introduction of the coordinated clause included to match sentence lengths.

Before examining the ability of LMs to predict reading times, we first report a series of control tests to assess the quality of the collected data, where we observe that human reading times follow the patterns described by sentence processing theories. We apply a Region of Practical Equivalence (ROPE) analysis, comparing RTs between different sentence groups and considering differences smaller than \num{0.1}~SD of the RT distribution (\SI{35}{\milli\second} in our data) as negligible. This helps us determine whether the observed differences are meaningful, rather than merely detectable due to our large sample size.

\paragraph{Structure drives difficulty beyond linear distance.}  We first observe the effect of dependency length on verb reading times, collapsing across clause types. Reading times increase from 0-word ($M=\SI{490}{\milli\second}$) to 4-word dependencies ($M=\SI{544}{\milli\second}$), and decrease slightly at 9 words ($M = \SI{528}{\milli\second}$). This pattern aligns partially with sentence processing theories, which argue that longer dependencies lead to higher processing costs, but already suggests that distance alone does not drive difficulty: the ROPE analysis confirms that both the 0- vs.\ 4-word ($d = -0.15$, \SI{95}{\percent} CI $[-0.19, -0.12]$) and 0- vs.\ 9-word contrasts ($d = -0.11$, \SI{95}{\percent} CI $[-0.15, -0.08]$) exceed the ROPE, while the 4- vs.\ 9-word contrast is negligible ($d = 0.05$, \SI{95}{\percent} CI $[0.02, 0.07]$). Collapsing across clause types, however, confounds distance with syntactic structure, so we turn to condition-level comparisons to investigate the impact of syntactic structure on processing difficulty.

\begin{figure}[t]
    \centering
    \includegraphics[width=1\linewidth]{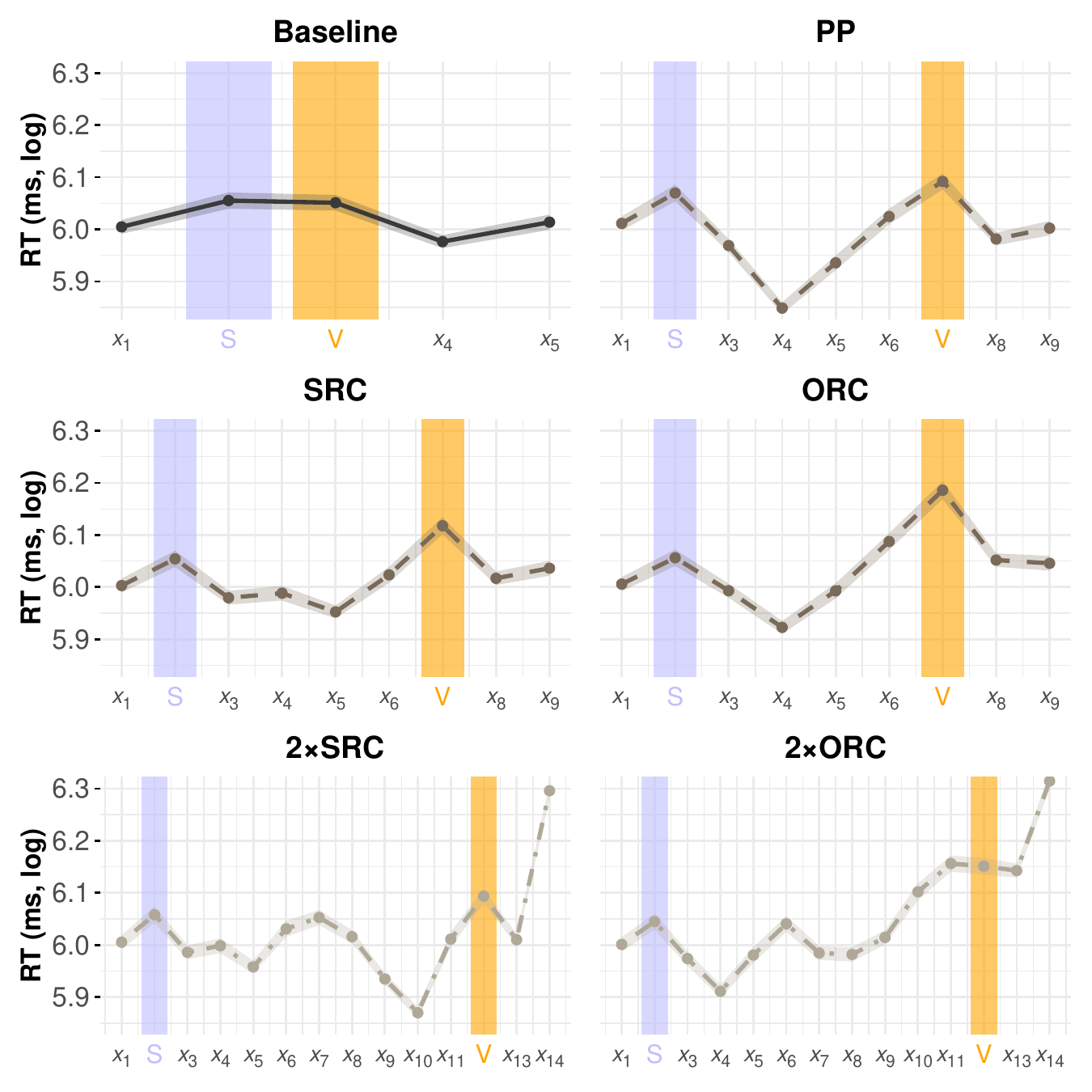}    
    \caption{Average log-transformed RTs by condition with \SI{95}{\percent} CIs (shaded ribbon). Curve colour indicates dependency length between \textcolor{blue!45}{\textbf{subject noun (S)}} and \textcolor{orange!55}{\textbf{main verb (V)}}: dark grey = 0-word; medium grey = 4-word; light grey = 9-word.}
    \label{fig:rt_by_position} 
\end{figure}

Figure~\ref{fig:rt_by_position} displays average log-transformed RTs at each word position for all six conditions and shows a consistent spike at the main verb across all conditions\footnote{
% The more gradual rise in 9-word dependency curves likely reflects successive reanalyses before resolution at the main verb. 
The end-of-sentence rise in the nested sentences (light gray curves) follows the expected SPR wrap-up effect; the other conditions end flat because their final word is the onset of a coordinated clause, not the sentence boundary.}. All conditions slow readers down relative to baseline, but to varying degrees (Table~\ref{tab:rt_by_condition}). Crucially, at the same 4-word dependency length, the three conditions pattern differently: the PP produces a negligible effect relative to the ROPE ($d = -0.08$, \SI{95}{\percent} CI $[-0.12, -0.04]$), the SRC a small but meaningful one ($d = -0.15$, \SI{95}{\percent} CI $[-0.19, -0.12]$), and the ORC the largest one ($d = -0.19$, \SI{95}{\percent} CI $[-0.24, -0.15]$). That three conditions sharing identical dependency length produce graded slowdowns suggests that the structural source of the dependency contributes to processing difficulty independently of its surface length. At 9-word dependencies, the same ordering holds: 2$\times$SRC ($d = -0.11$, \SI{95}{\percent} CI $[-0.15, -0.08]$) is slower than baseline but faster than 2$\times$ORC ($d = -0.14$, \SI{95}{\percent} CI $[-0.18, -0.10]$). Comparing across embedding depths, the increase from single to double embedding is modest: SRC and 2$\times$SRC differ by $\Delta d = 0.04$, and ORC and 2$\times$ORC by $\Delta d = 0.05$, both negligible relative to the ROPE. Clause type thus appears to be a stronger determinant of processing cost than embedding depth -- and, in fact, the direction of this difference runs counter to expectation: doubly-embedded conditions are consistently \emph{faster} than their singly-embedded counterparts for both clause types, a reversal we return to in Section~\ref{sec:discussion}, a reversal already visible in Table \ref{tab:rt_by_condition}.

% Notably, this is not merely a byproduct of the ROPE threshold: doubly-embedded conditions are consistently faster than their singly-embedded counterparts for both clause types. This reversal is corroborated by an independent, filler-calibrated estimate of empirical integration cost (described in Subsection \ref{subsec:filler-calibrated-expe}): nested-object dependencies show a smaller effect than standard-object ones (63 ms vs. 93 ms), and nested-subject dependencies show a smaller effect than standard-subject ones (14 ms vs. 49 ms). We return to possible explanations in Section \ref{sec:discussion}.

Mixed-effects models confirm that dependency length contributes to reading times even after accounting for standard psycholinguistic confounds (i.e., word length and frequency). In the following inferential analyses, we log-transform reading times (base 2), as is standard in psycholinguistics, which reduces skewness, stabilizes variance across conditions, and facilitates comparisons across studies. All models predict log-transformed RTs at the main verb with random intercepts for participants, items, and presentation order, fitted using the \verb|lme4| package \citep{bates2010lme4}. A baseline model (M0) including word length and log-transformed lexical frequency from WikiText-103 \citep{merity2016pointer} was compared via likelihood ratio test to a model additionally including dependency length (M1)\footnote{$log(RT)\sim log(freq)+len+dep+(1|part.) + (1|item)$}. The extended model significantly improves fit $\chi^2(1)=21.6$, $p<.001$,\footnote{The likelihood ratio test compares two nested models by asking whether the gain in fit, measured as the drop in negative log-likelihood, analogous to a drop in training loss, is larger than what chance alone would produce. The $\chi^2$ distribution provides the null reference for that drop.} and all three predictors reach significance: word length ($\beta = 0.012$, $t=5.04$, $p<.001$), log frequency ($\beta = -0.009$, $t=-4.10$, $p<.001$), and dependency length ($\beta = 0.0065$, $t=4.73$, $p<.001$). Crucially, the dependency length effect corresponds to an increase of approximately \SI{3}{\milli\second} per additional intervening word, a stable estimate in our large sample collected outside controlled laboratory conditions.

\begin{figure}[t]
    \centering
    \includegraphics[width=1\linewidth]{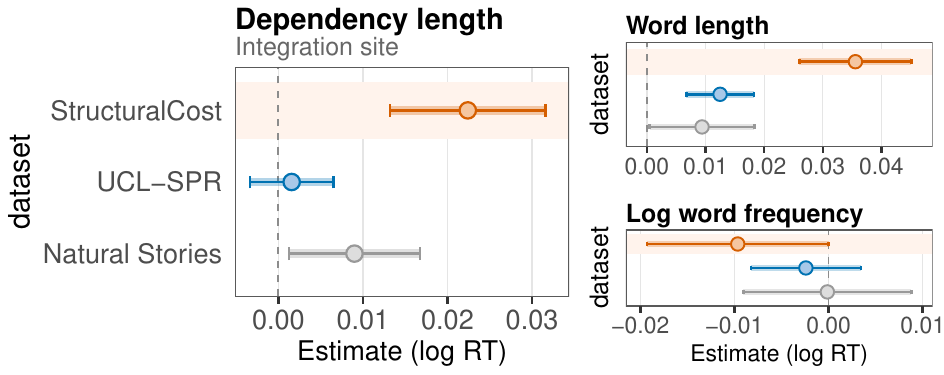}
    \caption{Fixed-effect estimates from M1 fitted separately to Natural Stories, UCL-SPR, and \benchname\ (at the integration site). All predictors are $z$-scored; estimates are on the log-RT scale. Error bars are \SI{95}{\percent} CIs.}
  \label{fig:combined_coef}
\end{figure}

\paragraph{Comparison with other datasets.}
To situate \benchname against related resources with less control, we fit the same model M1 to Natural Stories \cite{futrell2021natural} and UCL-SPR \cite{frank2013reading} at the integration site of the main verb and report the resulting fixed-effect coefficients in Figure~\ref{fig:combined_coef} (implementation details in Appendix~\ref{app:comparison_datasets}). The dependency length effect decreases at the main verb in both naturalistic corpora, while it increases in \benchname. Back-transforming from an example \SI{500}{\milli\second} baseline, this corresponds to a $\sim$1--5ms difference per SD of dependency length in naturalistic datasets versus $\sim$ \SI{11}{\milli\second} in \benchname~-- a difference our controlled design can isolate because dependency length is varied orthogonally to lexical properties. While word frequency effects are relatively similar across datasets, word length is a stronger RT predictor in \benchname than in the naturalistic corpora, which is expected given that our stimuli contain longer verbs by design (7.5 characters on average vs.\ 5 in the naturalistic datasets). These comparisons suggest that \benchname complements naturalistic corpora by providing experimental control over dependency length effects.
% Finally, word frequency shows a reliable slowing effect globally in \benchname, but this effect lowers drastically at the main verb, suggesting that lexical frequency influences incremental processing more broadly than at the main predicate specifically -- an observation that does not generalize to other corpora. Whether this reflects a property of the stimuli, the controlled design, or the reading task itself is left for future work.
% longer dependencies tend to involve rarer, longer words 
% if a typical baseline log RT is around ln(500ms) $\approx$ 6.21, then an estimate of $\seq$0.001 (UCL) and $\seq$0.009 (Natural Stories) corresponds to e$^$(6.21+0.005) $\approx$ 502ms so 2ms increase

\section{Probing Surprisal as a Cognitive Proxy}
We evaluate the extent to which LM surprisal captures the graded cost of integrating a subject with its verb across an intervening dependency as opposed to surface-level predictability. This requires dissociating two levels of alignment, which we term \textit{global} and \textit{local}.

\textit{Global} alignment asks whether surprisal correlates with RTs across all word positions. 
% Because it averages over every content word, function word, and phrase boundary, it captures a broad signal: whether the model finds roughly the same words hard to predict as humans find hard to read. 
This is a meaningful and well-established result \citep{smith2013effect}, but a weak test of cognitive plausibility: a model that has learned general word frequency and co-occurrence statistics will score well here, because most RT variance reflects lexical predictability rather than structural cost. Global alignment is therefore a necessary baseline but is not enough to model human processing: averaging over all 
positions dilutes the structural signal present at the main verb, which we evaluate separately as local alignment.

\textit{Local} alignment measures the same correlation restricted to a theoretically motivated position: the main verb, where the subject-verb dependency is resolved and humans show graded difficulty as a function of dependency length and syntactic structure. This is a much stricter test: fitting RTs well requires assigning higher surprisal when the dependency is harder to integrate, a sensitivity to structural cost that goes beyond surface statistics. A model that passes the global test but fails the local one has learned lexical predictability without encoding the memory-based integration cost that drives human processing difficulty.

% We provide a set of baseline experiments that illustrate potential usages of the dataset.

\subsection{Models and Setup}

\paragraph{Language models.} We selected models spanning a memory constraint spectrum: two KenLM $n$-gram models \cite{heafield2011kenlm}\footnote{Trained on WikiText-103, tokenized with \textsc{gpt-2} and \textsc{pythia} tokenizers.} as a Markovian baseline with a hard local window of 5 tokens; three \textsc{gpt-2} models (small, large, XL; \citealp{radford2019language}) and three \textsc{pythia} models (70M, 1.4B, 12B; \citealp{pmlr-v202-biderman23a}) as representatives of the currently dominant transformer architecture with unconstrained self-attention; and three \textsc{mamba} models (130M, 1.4B, 2.8B; \citealp{gu2024mambalineartimesequencemodeling}) implementing selective state-space compression as a proxy for recency-biased memory and incremental recurrent processing at scale. Within each family, size variation tests whether scaling improves behavioral fit while controlling for training data and vocabulary.

\paragraph{Surprisal extraction.} Word-level surprisal was extracted for every sentence, aggregating subword tokens following \citet{pimentel-meister-2024-compute}\footnote{Using the implementation of \citet{wisniewski2024probabilityword}.}.
Preprocessing details are given in Appendix~\ref{app:preprocessing}.

% For downstream analyses, use delta surprisal (relative to the baseline condition) to compare complex structures against a matched simple baseline. This way we control for lexical frequency at the cirtical word. Removes variance related to specific verb’s lexical freq: whatever makes “leaves” hard to predict in general is subtracted out, leaving only the structural contribution.
% Same for RTs? Has to be computed on averaged RTs per item (participants never see more than 1 sentence per set)

\paragraph{Statistical analyses}
We assess alignment between model surprisal and human RTs using two complementary methods. The first captures broad co-variation: for each word position, we compute Spearman correlations between model surprisal and mean RT across participants, evaluated both \textit{globally} (all positions) and \textit{locally} (main verb only). The second tests whether surprisal contributes independently of known low-level predictors: we fit a sequence of nested linear mixed-effects models~\citep{bates2010lme4} predicting log RT for each word read by each participant, adding predictors incrementally to isolate the unique contribution of surprisal and dependency length:

{\small
\begin{align*}
\textbf{M0}: \ & \text{log RT} \sim \text{log freq} + \text{length} + (1|\text{part.}) + (1|\text{item})\\
\textbf{M1}:\ & \cdots + \text{surprisal}\\
\textbf{M2}:\ & \cdots + \text{dependency length}
\end{align*}
}
Model comparison via likelihood ratio tests quantifies whether each added predictor genuinely improves fit. Note that in the global model, dependency length is a per-sentence condition-level predictor tested as a general robustness check across all word positions; it is only at the local (main-verb) level that dependency length functions as a theoretically motivated integration-cost predictor, since this is where the subject-verb dependency is resolved.

%
% Models are compared via likelihood ratio tests; the key quantity is $\Delta R^2$ from M0 to M1.

\subsection{Results}
\label{subsec:res}
\begin{figure}[htbp]
    \centering
    \includegraphics[width=1\linewidth]{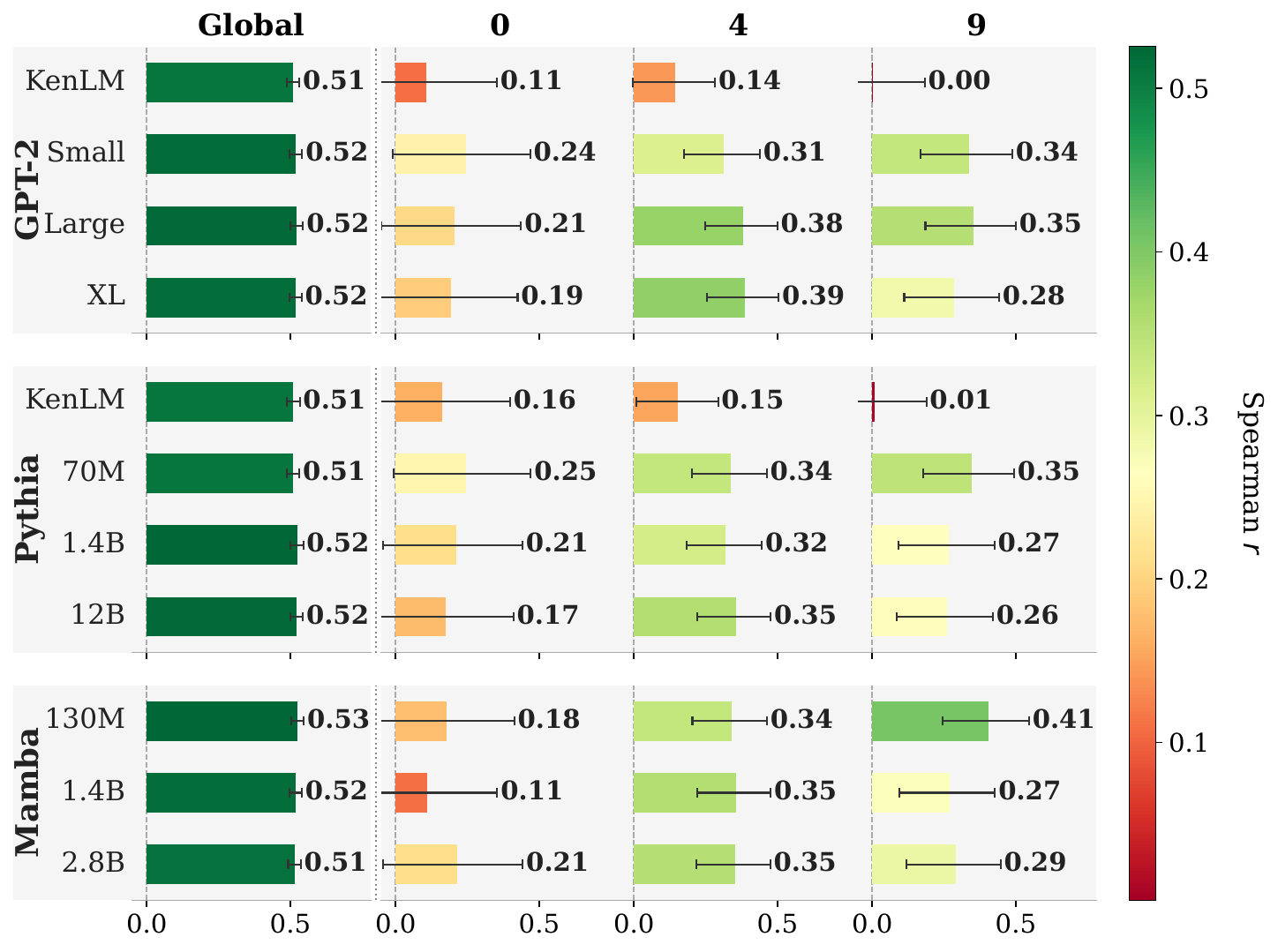}
    \caption{Spearman correlations between model surprisal and reading times, by dependency length (0, 4, 9 words) and fit level: all words (left) vs.\ verb only (right). Error bars show 95\% CIs. See Appendix~\ref{app:correlation} for the table.}
    \label{fig:correlation_heatmap}
\end{figure}

\paragraph{Global alignment: surprisal as a general RT predictor.}
We first establish whether LM surprisal predicts human RTs across all word positions, replicating prior work and setting a baseline against which local alignment can be assessed. Spearman correlations with word-level RTs are moderate and strikingly uniform: all models cluster at $r=0.51$--$0.53$ with tight CIs ($\pm .02$) far from zero (Figure~\ref{fig:correlation_heatmap}, left panel), regardless of architecture or size. The mixed-effects analysis confirms this picture: adding surprisal to the baseline model improves fit for all neural LMs ($p<.001$, $\chi^2(1)>846$; Table~\ref{tab:lmer_main}), indicating that surprisal captures variance in RTs beyond lexical frequency and word length. Further adding dependency length yields additional improvement ($p<.001$), suggesting that structural information contributes independently of surprisal. Global fit is thus robust, but as argued above, not sufficient: whether models encode the specific structural cost of subject-verb integration is the question we turn to next.

\begin{table}[t]
\centering
\footnotesize
\setlength{\tabcolsep}{2.5pt}
\begin{tabular}{lllll}
\toprule
& \multicolumn{2}{c}{M0$\to$M1} & \multicolumn{2}{c}{M1$\to$M2} \\
\cmidrule(lr){2-3}\cmidrule(lr){4-5}
Model & Global & Local & Global & Local \\
\midrule
\multicolumn{5}{l}{\textit{$N$-gram baselines}} \\
KenLM (GPT-2)  & $940^{***}$ & $0.02$      & $44^{***}$ & $20.5^{***}$ \\
KenLM (Pythia) & $993^{***}$ & $0.08$      & $44^{***}$ & $20.6^{***}$ \\
\multicolumn{5}{l}{\textit{GPT-2}} \\
Small          & $1082^{***}$ & $27.8^{***}$ & $33^{***}$ & $8.0^{**}$ \\
Large          & $1257^{***}$ & $43.4^{***}$ & $29^{***}$ & $5.0^{*}$ \\
XL             & $1010^{***}$ & $31.7^{***}$ & $31^{***}$ & $8.1^{**}$ \\
\multicolumn{5}{l}{\textit{Pythia}} \\
70M            & $930^{***}$  & $25.3^{***}$ & $42^{***}$ & $10.6^{***}$ \\
1.4B           & $1165^{***}$ & $20.8^{***}$ & $29^{***}$ & $9.4^{**}$ \\
12B            & $1005^{***}$ & $20.3^{***}$ & $29^{***}$ & $9.2^{**}$ \\
\multicolumn{5}{l}{\textit{Mamba}} \\
130M           & $1045^{***}$ & $23.5^{***}$ & $32^{***}$ & $9.7^{**}$ \\
1.4B           & $900^{***}$  & $18.2^{***}$ & $31^{***}$ & $11.3^{***}$ \\
2.8B           & $846^{***}$  & $18.7^{***}$ & $31^{***}$ & $9.4^{**}$ \\
\bottomrule
\end{tabular}
\caption{$\chi^2(1)$ from likelihood ratio tests, globally and locally 
at the main verb. M0: baseline (word log frequency + word length); M1: + surprisal; 
M2: + dependency length. $^{*}p<.05$, $^{**}p<.01$, $^{***}p<.001$. Global here tests dependency length as a sentence-level covariate across all positions; Local restricts to the integration site.}
\label{tab:lmer_main}
\end{table}

\begin{table*}[t]
\centering
\small
\begin{tabular}{lrrrrr}
\toprule
\textbf{Condition} & \textbf{Observed EOI (ms)} & \multicolumn{4}{c}{\textbf{Predicted EOI (ms)}} \\
\cmidrule(lr){3-6}
 & See Table \ref{tab:rt_by_condition} & KenLM (GPT2) & Mamba-130M & Pythia-1.4B & Pythia-70M \\
\midrule
PP              & 22 & 1.04 & 0.10  & $-$0.07 & $-$1.84 \\
SRC             & 49 & 0.02 & $-$1.18 & $-$1.25 & $-$2.13 \\
ORC             & 93 & 1.99 & 0.10  & 0.03  & $-$1.42 \\
2$\times$SRC    & 14 & $-$0.21 & $-$1.49 & $-$1.54 & $-$3.20 \\
2$\times$ORC    & 63 & 1.69 & 0.31  & 0.41  & $-$0.89 \\
\bottomrule
\end{tabular}
\caption{Predicted versus observed EOIs (relative to Baseline) at the main verb. Predicted values apply spillover-adjusted conversion factors, estimated on Natural Stories, to each model's surprisal on \benchname critical items; negative values indicate a predicted speedup. Models without a significant conversion factor are omitted.}
\label{tab:predicted_vs_empirical}
\end{table*}

\paragraph{Local alignment: surprisal at the integration site.}
% Cognitive plausibility requires more than global alignment. Humans show graded processing difficulty at the main verb as a function of dependency length -- this is what our dataset isolates. We now ask whether model surprisal captures this specific cost.
 
Restricting the analysis to the main verb reveals a consistent and informative drop (Figure~\ref{fig:correlation_heatmap}): at dependency length~0, no model yields reliable correlations, with all CIs including zero -- likely because the main verb follows immediately after the subject, leaving minimal left context for surprisal to track. At dependency length 4, correlations fall from $r=0.51$--$0.53$ globally to $r=0.31$--$0.39$ at the verb for neural models, and collapse to $r=0.14$--$0.15$ for $n$-gram models, whose CIs include zero ($[-.01, .29]$). Neural models show significant surprisal effects at dependency length~4 (all $p < 10^{-5}$, $\chi^2(1) = 18$--$43$), with CIs entirely above zero, and, when comparing model surprisal at the main verb itself across conditions matched for linear distance, reproduce the syntactic ordering \textsc{pp} $<$ \textsc{src} $<$ \textsc{orc} at fixed linear distance (all $p < .001$), confirming sensitivity to hierarchical structure beyond raw word count. Dependency length contributes further beyond surprisal (all $p < .05$, $\chi^2(1) = 5$--$11$), indicating that structural distance is not fully captured by surprisal alone. At dependency length~9, neural models hold up but unevenly: most settle at $r = 0.26$--$0.35$ with wider CIs reflecting reduced power, the \textsc{src} $<$ \textsc{orc} verb-surprisal contrast attenuates, and no model approaches the sharp \SI{70}{\milli\second} \textsc{orc} penalty humans incur at fixed linear distance (PP vs. ORC in Table~\ref{tab:rt_by_condition}), an indication that the failure is qualitative, not merely quantitative. $N$-gram models fail entirely: with the subject noun beyond the 5-gram window at dependency \num{4} and \num{9}, surprisal adds nothing over baseline predictors at any dependency length ($p > 0.5$, $\chi^2(1) < 0.1$). This global-to-local reduction is the central finding of the present work: models that appeared equivalent under global alignment diverge sharply when tested at the integration site.
 
Crucially, scaling within families has no consistent benefit: larger models do not fit human integration profiles better, and the largest variants lose correlation points relative to their smaller counterparts. This suggests the gap is not a matter of capacity but a structural one (e.g., inductive bias or training objective). Consistent with this, \textsc{mamba-130M} reaches $r = 0.41$ at dependency length~9, the only model whose CI does not overlap with any 0-word dependency interval, approaching the global baseline and outperforming larger attention-based models. This suggests that memory-constrained architectures produce integration profiles closer to human behavior, though the advantage remains modest and requires further investigation. Reader profile analyses (Appendix~\ref{app:reader_profile}) point to a further limitation independent of architecture or scale: surprisingly, LMs fit slow and medium readers better than fast ones, who display high working-memory readers and are simultaneously faster and more accurate. The fast readers being the hardest to model suggests efficient integration strategies remain outside what surprisal captures.

\subsection{Comparison of predicted and observed RTs}
\label{subsec:filler-calibrated-expe}

If surprisal captured the full integration cost of the dependency, a surprisal-to-RT conversion factor estimated independently should predict RT effects of the observed magnitude. Following
\citet{van2021single} and \citet{huang2024large}\footnote{Originally developed to test whether surprisal explains syntactic disambiguation difficulty.}, we test this by estimating, for each model, a conversion factor (ms per bit of surprisal) on an independent corpus (Natural Stories, \citealp{futrell2021natural}), regressing RT on surprisal at the current and three preceding words (spillover), word length, and log-frequency\footnote{Pythia-12B and KenLM\_Pythia were excluded due to technical constraints.}. Only four models yielded a significant spillover coefficient at $p<.01$ (Mamba-130M, Pythia-70M, Pythia-1.4B, KenLM/GPT-2) and were retained as conversion factors; we applied their conversion factors to surprisal on \benchname{} critical items to obtain predicted Effects of Interest (EOIs)\footnote{Predicted RT minus predicted Baseline RT.}.
Table~\ref{tab:predicted_vs_empirical} compares predicted and observed EOIs. Predicted effects were between one and three orders of magnitude too small across all four models: for KenLM-GPT2, predicted EOIs ranged \qtyrange{0.02}{1.99}{\milli\second} against observed effects of \qtyrange{14}{93}{\milli\second}, while Pythia-70M predicted a speedup (negative EOI) in every condition. No model recovered the correct direction of the effect in the 2$\times$SRC condition, where predictions were near-zero or negative despite a small observed slowdown. Even calibrated on an independent corpus and estimated via a completely separate method, surprisal did not reproduce the scale of the observed integration cost.

\section{Discussion}
\label{sec:discussion}

Our results both replicate and qualify prior work \citep{futrell2021natural, oh2023transformer, shain2024large}: global surprisal-RT alignment is robust but dominated by lexical frequency and 
word length, explaining why correlations cluster at $r \approx 0.5$ regardless of parameter count, training data, or memory architecture. Any model with basic word statistics will track these signals, which is precisely why global alignment is a weak diagnostic for cognitive plausibility that does not generalize to structural difficulty. Neural models significantly improve fit at the main verb over $n$-gram baselines and partially mirror the syntactic ordering of conditions, yet they underestimate how processing cost scales with dependency length. The gap persists across architectures and sizes, pointing to a specific limitation: current models appear to capture the predictive component of processing but not the memory-based integration cost that working memory imposes at structural dependencies. The one partial exception is \textsc{mamba}-130M, which achieves the highest verb-level correlations at long dependencies, consistent with the hypothesis that selective state-space compression introduces a recency bias that partially mimics distance-dependent memory decay. 
% \textsc{gpt-2} large and XL show similar promise, in line with prior work on their candidacy as cognitive models. 
Whether these advantages reflect genuine architectural sensitivity or confounds in training data and tokenization remains to be established.
An unexpected pattern emerges at double embedding: 2$\times$SRC and 2$\times$ORC are consistently read faster than their singly-embedded counterparts.
%(2$\times$ORC: \num{63}ms vs. ORC: \num{93}ms; 2$\times$SRC: \num{14}ms vs. SRC: \num{49}ms). 
No model predicts this reversal. One possible explanation is within-sentence syntactic adaptation: by the second embedding, readers have already processed the relevant structure once earlier in the same sentence, potentially lowering marginal integration cost even as cumulative memory load increases. We flag this as an open question for future work rather than a settled mechanism.
Together, these findings motivate a reorientation away from global evaluation and toward controlled datasets that isolate memory-sensitive positions. On the modeling side, they favor architectural interventions that introduce explicit memory constraints such as limited context \citep{kuribayashi2022context}, recency-biased attention \citep{devarda2024locally, clark-etal-2025-linear}, or capacity-constrained recurrence \citep{timkey2023language} over further scaling. \benchname~provides the evaluation framework needed to test such interventions directly at the integration site.

% They also have a superhuman memory capacity (e.g., \cite{wilcox_etal_2025}). In terms of short term memory access, they have a decay-free access to huge context window (thousands of tokens at ones vs ~4 information chunks for an healthy human adult). In terms of long-term memory, they are trained on massive corpora (up to 15T words vs ~100M for a child), favoring memorization over efficient generalization (e.g., \citealp{oh2025model}).

\section{Conclusions}

We introduced \benchname, a controlled self-paced reading dataset isolating the processing cost of dependency length from lexical confounds. Human reading times at the main verb increase with dependency length, driven by syntactic embedding beyond linear distance, confirming that the dataset captures memory-based difficulty.
Using \benchname, we evaluated LMs spanning $n$-gram, SSM, and transformer architectures. Two findings emerge consistently. First, global surprisal-RT alignment is architecturally uninformative: models cluster at similar correlations regardless of architecture or scale, reflecting the dominance of lexical frequency and word length in RT variance. Second, this global alignment does not extend to structural difficulty: neural models partially mirror the graded difficulty profile at the main verb, yet underestimate the integration cost humans incur, and this gap persists across architectures and sizes.
These findings suggest that global surprisal-RT alignment is not a sufficient criterion for cognitive plausibility. Current models seem to capture the predictive component of human processing but not the memory-based integration cost. \benchname~provides the controlled evaluation data needed to make progress on this gap.

\section*{Limitations}
\benchname~prioritizes internal validity at the cost of ecological generalizability: sentences are controlled constructions in English only, collected online from native speakers, limiting cross-linguistic and cross-population conclusions. The online setting also means the reading environment and equipment were not controlled. 
A denser condition gradient (e.g., a PP+SRC combination) would further help dissociate linear and hierarchical contributions, and would clarify the embedding-depth reversal noted in Section \ref{sec:discussion}; we leave this to future work.
Working memory capacity scores, though collected, are not yet fully leveraged here; a fuller individual-differences analysis is a natural extension.
The 0.7 comprehension accuracy cutoff is conventional; participants just below this threshold may have been engaged readers. Due to an implementation error, the dataset only contains each participant's overall comprehension accuracy, not their per-item comprehension results; the latter would have allowed a finer-grained comparison between reading times and comprehension accuracy at the item level.
On the modeling side, LM surprisal captures predictive processing cost but not the integration cost arising from working memory constraints: a limitation that motivates the dataset rather than undermining it but that questions nonetheless the reliability  of surprisal as a complete proxy for human processing difficulty. Finally, we evaluate a representative but non-exhaustive set of architectures; other models may behave differently, and their evaluation on \benchname~remains open.

\section*{Ethical considerations}
This study has required an interaction with humans. As such, it has been reviewed and approved by an Institutional Review Board (IRB No. 00012024-128).

\section*{Acknowledgments}
This research has been financially supported by the Agence Nationale de la Recherche (project COMPO, ANR-23-CE23-0031-01). We also want to thank the EMNLP reviewers for their constructive suggestions, as well as those from participants of the \textit{Computational Psycholinguistics Meeting 2025} where the dataset had been initially presented.

% % The experiments in this work were carried out on Magi, the experimental platform of University of Sorbonne Paris North (USPN) dedicated to research. This platform offers researchers at the institution High-Performance Computing (HPC), cloud and storage services.
% Use of AI: If you used generative AI in acceptable ways, e.g. coding or writing assistance, please disclose it in either the acknowledgements or the checklist itself.

% Bibliography entries for the entire Anthology, followed by custom entries
%\bibliography{anthology,custom}
% Custom bibliography entries only
\bibliography{anthology, custom}

\begin{table*}[htpb]
\centering
\small
\renewcommand{\arraystretch}{1.3}
\setlength{\tabcolsep}{3.8pt}
\begin{tabular}{llllllcccc}
\hline
\textbf{Dataset} &  \textbf{} & \textbf{Method} & \textbf{Uniq. Sent.} & \textbf{Subj.} & \textbf{Subj./Item} & \textbf{C1} & \textbf{C2} & \textbf{C3} & \textbf{C4} \\
\hline
\multicolumn{10}{l}{\textit{Naturalistic datasets}} \\
\hline
Natural Stories  & \citealp{futrell2021natural}    & SPR      & 485 & 181 & $\sim$100    & \xmark & \xmark & \cmark & \cmark \\
Dundee        & \citealp{kennedy2003dundee}      & ET       & 2,368 & 10 & 10    & \xmark & \xmark & \cmark & \xmark  \\
Provo          & \citealp{luke2018provo}       & ET       & $\sim$138 & 84 & $\sim$8    & \xmark & \xmark & \both & \xmark \\
UCL Corpus      & \citealp{frank2013reading}      & SPR + ET & 361 + 205 & 117 + 43 & *   & \cmark & \xmark & \cmark & * \\
MECO           & \citealp{kuperman2024MECO}       & ET       & 104 & 50 & *    & \xmark & \xmark & \both & * \\
GECO          & \citealp{cop2017geco}        & ET       & 5,300 & 14 & 14    & \xmark & \xmark & \cmark & \xmark  \\
ZuCo          & \citealp{hollenstein2018zuco}        & ET + EEG & 1,107 & 12 & 12    & \both & \xmark & \cmark & \xmark \\
\hline
\multicolumn{10}{l}{\textit{Controlled datasets}} \\
\hline
Sent. Complexity    & \citealp{grodner2005consequences} & SPR    & 16 + 30 & 42+49 & 21 \& 49    & \cmark & \cmark & \xmark & \xmark\\
Syn. Ambiguity     & \citealp{huang2024large}   & SPR    & 18-24 & 2,000 & 220-440   & \cmark & \cmark & \xmark & \cmark \\
\hline
\multicolumn{10}{l}{\textit{This work}} \\
\hline
\textbf{\benchname} & & \textbf{SPR} & \textbf{450} & \textbf{475} & \textbf{80} & \cmark & \cmark & \cmark & \cmark \\
\hline
\end{tabular}
\caption{Comparison of reading time datasets against four criteria required to study LLM replication of human processing difficulty (non-exhaustive list).
\textbf{C1}: isolated sentences;
\textbf{C2}: systematic dependency length manipulation;
\textbf{C3}: sufficient items for model evaluation; 
\textbf{C4}: sufficient participants per item.
ET = eye-tracking; SPR = self-paced reading.
\cmark~= satisfied; \xmark~= not satisfied; \both~= partially satisfied; *~= information not found.}
\label{tab:dataset-comparison}
\end{table*}

\appendix
\section*{Appendices}
\section{Dataset comparison}
\label{app:dataset_comparison}

Table~\ref{tab:dataset-comparison} compares \benchname against existing reading time datasets across the four criteria we argue are necessary to study LLM replication of human processing difficulty.

\section{\benchname datasheet}
\label{app:appendix_datasheet}

\subsection{Design \& conditions}

\paragraph{Critical conditions.}
Each of 60 item sets contains 6 sentences sharing the same main subject-verb (NV) pair.
Conditions vary dependency length and syntactic structure as shown in Table~\ref{tab:stimuli}.

\paragraph{Control conditions.}
Control sentences isolate linear distance effects without clause embedding.
They do not share NV pairs with critical items and use only prepositional phrases
to increase distance (Table~\ref{tab:stimuli}).

\paragraph{Sentence construction constraints.}
All critical and control sentences were subject to the following constraints during generation and manual review:
\begin{itemize}
    \item All six sentences in a set share the same main subject and main verb, with intervening words held constant or maximally similar across conditions.
    \item Main verbs are never sentence-final to avoid spillover and end-of-sentence effect ambiguity.
    \item Embedded nouns match the number of the main subject to prevent agreement-attraction confounds (e.g., “The \textit{babysitters}$_{PL}$ who the \textit{children}$_{PL}$ loved prepare the diner”).
    \item Embedded clauses use past tense while main clauses use present tense.
    \item All sentences are 13--15 words; shorter conditions are padded with a post-critical coordinated clause to match length (see Table \ref{tab:stimuli}).
    \item Cloze-likely verbs were replaced to avoid predictability artifacts (e.g., “The baker bakes” or “The violinist plays” where “prepares” or “leaves” were preferred).
    \item Each sentence was judged semantically correct and plausible by a proficient speaker.
\end{itemize}    

\newpage
The revision pipeline involves: 
\begin{itemize}
    \item Proportion of sentence sets substantially modified or replaced: ~60\% for the critical sentences and ~90\% for the control sentences; \item Number of reviewers: 1 for the critical sentences and 2 for the control sentences; \item Total hours: $\pm 70$. 
\end{itemize}

\paragraph{Comprehension assessment.}
True/false comprehension questions followed more than half (56\%) of sentences to ensure active processing \citep{ferreira2019problem}. Each participant answered 30 questions on control sentences and 20 on critical ones, with an equal number of true and false answers, ensuring consistent reading strategy across sentence types. Participants received feedback via a progress bar showing their running accuracy. Questions were not created for doubly embedded conditions (2$\times$SRC and 2$\times$ORC), which are already taxing to process, and never targeted the main verb to avoid drawing attention to the critical region.

\paragraph{Items per participant.}
Table~\ref{tab:item_counts} summarizes the number of items seen per participant. The 30 critical 4-word dependency length items comprise 10 PP + 10 SRC + 10 ORC. The 20 critical 9-word dependency length items comprise 10 2$\times$SRC + 10 2$\times$ORC. 
\begin{table}[htpb]
\centering
\footnotesize
\begin{tabular}{@{}lcccc@{}}
\toprule
 & \textbf{Dep.\ 0} & \textbf{Dep.\ 4} & \textbf{Dep.\ 9} & \textbf{Total} \\
\midrule
Critical & 10 & 30 & 20 & \textbf{60} \\
Control   & 10 & 10 & 10 & \textbf{30} \\
\midrule
Total    & \textbf{20} & \textbf{40} & \textbf{30} & \textbf{90} \\
\bottomrule
\end{tabular}
\caption{Items per participant by dependency length.}
\label{tab:item_counts}
\end{table}

\subsection{Item counts \& list structure}
\paragraph{Latin square counterbalancing.}
Items were distributed across 6 lists using a Latin square design so that each participant saw exactly one sentence per item set, with condition assignment rotating across lists. Each list contains 60 critical items (1 per set), a 50\%/50\% singular/plural number balance, and 30 control sentences.

\subsection{Participants: recruitment and payment}
To avoid confounding structural difficulty with proficiency effects, we restrict our analysis to native speaker data exclusively. Participants were recruited via Prolific, restricted to self-reported native English speakers with U.S. nationality or longest residence in the U.S., and no diagnosed or suspected reading disorder. The experiment lasted approximately 30 minutes and participants were compensated £5 (£10/hour, approximately \$13.50/hour), consistent with Prolific's fair pay guidelines.

\subsection{Consent procedure}
Before the experiment began, participants were presented with a full information letter 
describing the study objectives, procedure, risks, and their rights (including the right to withdraw at any time without consequence). They provided explicit consent to participate in a study on reading times, acknowledging that no directly identifying data would be reported. Only consenting participants were admitted. Data was pseudonymized upon collection and anonymized before analysis. 

\subsection{Instructions given to participants}

\paragraph{General instructions.}
\begin{quote}
\textit{This experiment consists of two tasks: a reading comprehension task and a memory test. 
In the first part, you will read several sentences, one word at a time, and answer a 
comprehension question if one follows. In the second part, you will evaluate the correctness 
of simple math equations while memorizing letters which will appear in between. Please make 
sure you are in a quiet environment and have a stable internet connection before you begin.}
\end{quote}

\paragraph{Reading task instructions.}
\begin{quote}
\textit{You will now read a series of sentences and answer a simple question about some of 
them. The sentences will appear one word at a time. Press the SPACE bar to reveal each new 
word. You cannot go back to previous words, so read each sentence carefully and naturally at 
a pace that feels normal to you. Your goal is to read for comprehension in order to score 
80\% accuracy or higher on the comprehension questions.}
\end{quote}

\paragraph{Memory task instructions.}
\begin{quote}
\textit{In this task, you will do two things: evaluate the correctness of math equations and 
memorize the letters that appear after each item. Read each math equation aloud and press Y 
if the equation is correct, or N if it is incorrect. After a series of items, you will be 
asked to type in the list of letters in the order that you read them.}
\end{quote}

\paragraph{Debrief.}
\begin{quote}
\textit{In this study, we measured whether the length of dependencies between words in a 
sentence increases reading time. Some sentences were more complex and difficult to process — 
they were included to test the limits of human processing capacity. These measures will serve 
to constrain language models to process language more like humans.}
\end{quote}

% \begin{table*}[ht]
% \centering
% \footnotesize
% \begin{tabular}{lp{9cm}}
% \toprule
% \textbf{Parameter} & \textbf{Value / Description} \\
% \midrule
% Total recruited         & X \\
% Final $N$ (after exclusions) & 475 \\
% Age (mean $\pm$ SD)     & $44.7 \pm 12.5$ years (range: 20--70) \\
% Recruitment platform    & Prolific \\
% Nationality criterion   & U.S.\ nationality, or longest residence in the U.S. \\
% Language criterion      & Self-reported native English speaker \\
% Exclusion 1             & Self-reported reading disorder $\to$ excluded before participation \\
% Exclusion 2             & Comprehension accuracy $< 0.70$ $\to$ excluded post-hoc ($N = $ [X]) \\
% Items per participant   & 90 sentences (60 critical + 30 controls), one randomly assigned list \\
% Observations (final)    & 28{,}500 trial-level reading time observations \\
% \bottomrule
% \end{tabular}
% \caption{Participant summary.}
% \label{tab:participants}
% \end{table*}

\subsection{Data collection procedure}

\paragraph{Self-paced reading task.}
In self-paced reading, sentences are presented word by word, with each word appearing only when the participant presses a key; reading times are recorded as the time between keypresses. The task was implemented in jsPsych, adapted from \citet{vondermalsburg2015spr} for online 
administration via Prolific.

\paragraph{Working memory capacity.}
Working memory capacity (WMC) was assessed using an operation span task \citep{conway2005working}, in which participants solve arithmetic operations while maintaining sequences of letters in memory. Scores ranged from 0 to 6 (Table~\ref{tab:wm_distribution}), with concentration in mid-range values consistent with healthy adult populations. The task was adapted from \citet{vondermalsburg2015span}.

\begin{table}[htpb]
\centering
\footnotesize
\begin{tabular}{@{}ccc@{}}
\toprule
\textbf{OSpan score} & \textbf{$N$} & \textbf{\%} \\
\midrule
0 & 38  & 8  \\
1 & 0   & 0  \\
2 & 32  & 7  \\
3 & 84  & 18 \\
4 & 131 & 28 \\
5 & 116 & 24 \\
6 & 74  & 16 \\
\midrule
Total & 475 & 100.0 \\
\bottomrule
\end{tabular}
\caption{Distribution of operation-span (OSpan) scores across participants. Protocol: \citet{conway2005working}; implementation adapted from von der Malsburg's Python version.}
\label{tab:wm_distribution}
\end{table}

\subsection{Stimulus generation}

\paragraph{Generation pipeline.}
Critical sentences were generated using the LLaMA-3.2 API \citep{grattafiori2024llama}, prompting the model with a semantic domain (e.g., \textit{school}, \textit{health}, \textit{weather}) and grammatical number (singular/plural) , and requesting all six structural variants simultaneously with a shared subject-verb pair and maximal lexical overlap. 
% Full prompts are provided in Appendix~\ref{app:prompts}. 
Control sentences were generated one set at a time using Perplexity.ai \citep{perplexity2024}, specifying the PP-only constraint, three dependency lengths (0, 4, 9 words), and maximal lexical overlap. All items then underwent manual review for plausibility, verb predictability, number agreement, tense consistency, and length matching (13-15 words).

\section{Preprocessing}
\label{app:preprocessing}

The first and last word of each sentence are excluded from all analyses: SPR paradigms produce elevated RTs at sentence boundaries that reflect task mechanics rather than linguistic processing. As a byproduct, this removes the beginning-of-sequence token whose surprisal is undefined or uninformative across LMs. These exclusions are applied uniformly to RT and surprisal files.

Reader profiles are computed on trimmed data: each participant's mean RT is calculated across all retained word positions, and participants are partitioned into fast, medium, and slow profiles at the 33rd and 66th percentiles. All predictors are z-scored to make estimates comparable across datasets.

\section{Dependency parsing}
\label{app:comparison_datasets}

Dependency length is operationalized as the linear distance in tokens between the main verb and its subject, extracted automatically via UDPipe \cite{straka2016udpipe} using the English Universal Dependencies model \cite{nivre2016universal} for UCL corpus. Natural Stories already provide parse files.

\section{Reader profile alignment}
\label{app:reader_profile}

Human variability is non-trivial and theoretically informative. Thus we compute reader profiles on the previously trimmed data: each participant's mean RT is calculated across all retained word positions, and participants are partitioned into fast, medium, and slow profiles based on the 33rd and 66th percentiles of this distribution. These profiles are used throughout as secondary diagnostics.

We observe an asymmetry concerning which readers LMs approximate most closely. Contrary to the prediction that surprisal, a prediction-based signal, would align with fast, anticipation-driven readers, the opposite holds: fast readers show the lowest verb-level fit across all models ($r=0.15$-$0.22$ for neural models; near zero for KenLM), while medium and slow readers show consistently higher fit ($r= 0.36$-$0.40$). Cognitive profiling suggests this is not an artifact of inattention: high working-memory readers are simultaneously faster and more accurate, yet the ones LMs fit least well. One interpretation is that efficient readers deploy an integration strategy that bypasses the retrieval operations surprisal tracks, a dimension of human processing that current LMs, regardless of architecture or scale, do not capture.

% \begin{figure}[ht]
%     \centering
%     \includegraphics[width=1\linewidth]{images/wmc_reading_profile2.png}
%     \caption{Enter Caption}
%     \label{fig:wmc_reader_profile}
% \end{figure}

% \begin{figure}[ht]
%     \centering
%     \includegraphics[width=1\linewidth]{images/surprisal_RT.png}
%     \caption{Enter Caption}
%     \label{fig:surprisal_rt}
% \end{figure}

% \section{Surprisal--RT correlation}
% \label{sec:surprisal_rt_corr}

% \begin{figure}[ht]
%     \centering
%     \includegraphics[width=1\linewidth]{images/heatmap_global_local_Fast.pdf}
%     \caption{Enter Caption}
%     \label{fig:heatmap_fast}
% \end{figure}

% \begin{figure}[ht]
%     \centering
%     \includegraphics[width=1\linewidth]{images/heatmap_global_local_Med.pdf}
%     \caption{Enter Caption}
%     \label{fig:heatmap_med}
% \end{figure}

% \begin{figure}[ht]
%     \centering
%     \includegraphics[width=1\linewidth]{images/heatmap_global_local_Slow.pdf}
%     \caption{Enter Caption}
%     \label{fig:heatmap_slow}
% \end{figure}

\section{Surprisal-RT correlation}
\label{app:correlation}
Spearman is used instead of Pearson as the relation between surprisal and RT is not fully linear. Table~\ref{tab:spearman} shows Spearman correlations between model surprisal and mean per-word RT, globally and locally at the main verb by dependency length.

\begin{table}[htpb]
\centering
\small
\setlength{\tabcolsep}{2.5pt}
\begin{tabular}{lcccc}
\toprule
& \multicolumn{4}{c}{Spearman $r$ $[95\%$ CI$]$} \\
\cmidrule(lr){2-5}
Model & Global & Dep=0 & Dep=4 & Dep=9 \\
\midrule
\multicolumn{5}{l}{\textit{$N$-gram}} \\
KenLM\_G   & .51 \scriptsize{[.49,.53]} & .11 \scriptsize{[-.14,.35]} & .14 \scriptsize{[-.01,.28]} & .00 \scriptsize{[-.18,.19]} \\
KenLM\_P  & .51 \scriptsize{[.49,.53]} & .16 \scriptsize{[-.09,.39]} & .15 \scriptsize{[.01,.29]}  & .01 \scriptsize{[-.17,.19]} \\
\midrule
\multicolumn{5}{l}{\textit{GPT-2}} \\
Small           & .52 \scriptsize{[.50,.54]} & .24 \scriptsize{[-.02,.47]} & .31 \scriptsize{[.17,.45]}  & .34 \scriptsize{[.16,.51]} \\
Large           & .52 \scriptsize{[.50,.55]} & .21 \scriptsize{[-.05,.44]} & .38 \scriptsize{[.25,.51]}  & .35 \scriptsize{[.18,.52]} \\
XL              & .52 \scriptsize{[.50,.54]} & .19 \scriptsize{[-.05,.42]} & .39 \scriptsize{[.26,.51]}  & .28 \scriptsize{[.12,.45]} \\
\midrule
\multicolumn{5}{l}{\textit{Pythia}} \\
70M             & .51 \scriptsize{[.49,.53]} & .25 \scriptsize{[-.00,.50]} & .34 \scriptsize{[.19,.47]}  & .35 \scriptsize{[.17,.51]} \\
1.4B            & .52 \scriptsize{[.50,.55]} & .21 \scriptsize{[-.05,.45]} & .32 \scriptsize{[.18,.46]}  & .27 \scriptsize{[.08,.44]} \\
12B             & .52 \scriptsize{[.50,.55]} & .17 \scriptsize{[-.09,.41]} & .35 \scriptsize{[.21,.48]}  & .26 \scriptsize{[.08,.43]} \\
\midrule
\multicolumn{5}{l}{\textit{Mamba}} \\
130M            & .53 \scriptsize{[.50,.55]} & .18 \scriptsize{[-.07,.42]} & .34 \scriptsize{[.21,.47]}  & \textbf{.41} \scriptsize{[.24,.57]} \\
1.4B            & .52 \scriptsize{[.50,.54]} & .11 \scriptsize{[-.14,.35]} & .35 \scriptsize{[.21,.48]}  & .27 \scriptsize{[.09,.44]} \\
2.8B            & .51 \scriptsize{[.49,.54]} & .21 \scriptsize{[-.04,.43]} & .35 \scriptsize{[.21,.49]}  & .29 \scriptsize{[.10,.46]} \\
\bottomrule
\end{tabular}
\caption{Spearman correlations between model surprisal and mean per-word RT, globally and locally at the main verb by dependency length. Bold marks the single best local score. CIs computed via bootstrap ($k=1000$).}
\label{tab:spearman}
\end{table}

% \section{Prompts}
% \label{app:prompts}
\end{document}